\documentclass{article}

\usepackage{geometry}
\usepackage[style=authoryear,backend=biber]{biblatex}
\usepackage{amsmath, amssymb, amsthm}
\usepackage{tcolorbox}

\usepackage{graphicx}
\usepackage{caption}
\usepackage{array}
\usepackage{booktabs, multirow}
\usepackage{float}
\usepackage{wrapfig}
\usepackage{subcaption}
\usepackage{microtype}
\usepackage[hidelinks]{hyperref}

\makeatletter
\newcommand{\customlabel}[2]{%
	\protected@write \@auxout {}{\string \newlabel {#1}{{#2}{\thepage}{#2}{#1}{}} }%
	\hypertarget{#1}{}
}
\makeatother

\newcommand{\half}{\frac{1}{2}}
\newcommand{\fhat}{\hat{f}}

\begin{document}

\begin{center}
	{\Large Backpropagation through Time and Space:\\Learning Numerical Methods with Multi-Agent Reinforcement Learning}
\end{center}

\vspace{20pt}


\hspace{-38pt} 
\setlength{\tabcolsep}{3.4pt}
\begin{tabular}{c*{4}{>{\centering\arraybackslash}m{4.075cm}}}
	&
	\textbf{Elliot Way}
	
	Binghamton University
	&
	\textbf{Dheeraj S.K.~Kapilavai}
	
	GE Global Research
	&
	\textbf{Yiwei Fu}
	
	GE Global Research
	&
	\textbf{Lei Yu}
	
	Binghamton University
\end{tabular}

\vspace{20pt}

\begin{abstract}
	We introduce Backpropagation Through Time and Space (BPTTS), a method for training a recurrent spatio-temporal neural network, that is used in a homogeneous multi-agent reinforcement learning (MARL) setting to learn numerical methods for hyperbolic conservation laws. We treat the numerical schemes underlying partial differential equations (PDEs) as a Partially Observable Markov Game (POMG) in Reinforcement Learning (RL). Similar to numerical solvers, our agent acts at each discrete location of a computational space for efficient and generalizable learning. To learn higher-order spatial methods by acting on local states, the agent must discern how its actions at a given spatiotemporal location affect the future evolution of the state. The manifestation of this non-stationarity is addressed by BPTTS, which allows for the flow of gradients across both space and time. The learned numerical policies are comparable to the SOTA numerics in two settings, the Burgers' Equation and the Euler Equations, and generalize well to other simulation set-ups.
\end{abstract}

\section{Introduction}

Simulations of physical processes in multiphysics systems like gas turbines, weather, etc., involve several layers of numerical algorithm decision making that aims to march a current state to a future state. The design of the numerical scheme requires substantial effort by experts with understanding of the underlying physics and numerical methods to translate the partial differential equations (PDEs) into a computable model. Typically, simulation involves computing fluxes of conserved quantities in a discretized volume and time integrating to determine the next state.  This incremental computation of the physics in time is a sequential decision-making process akin to Reinforcement Learning (RL) problems.

Modeling PDEs numerically requires making decisions at every point across the physical space. This entails a wide space of continuous actions for RL, containing at best hundreds of dimensions, depending on how the space is discretized. Learning a policy for every action at once is intractable; instead, it is necessary for the RL agent to act on a local level, selecting locally applicable actions based on observations from its neighborhood like in a physics simulation. The learned policy is applied repeatedly across all locations, or equivalently, each discrete location is assigned an identical agent. RL with multiple identical agents is known as homogeneous multi-agent RL (MARL).
Learning with MARL is difficult because actions impact the state at other locations, and training must account for the confounding influence of distant actions. Consider a large volume where physics propagate from point $x$ to $y$. If the action at $x$ changes, the observations at $y$ change, even if the actions at $y$ are the same. At $y$, the environment appears non-stationary; the effect actions have on subsequent observations does not stay the same because observations also depend on actions at other locations.

Traditional RL algorithms learn from separated state-to-state transitions, but this is ineffective with multiple agents as it loses the context of actions at other times and locations. Our solution takes inspiration from \textcite{mordatch2017emergence}. They handle non-stationarity by constructing a recurrent neural network that encodes the actions of every agent for the entire episode, then optimize the network to maximize rewards. We adapt this solution to PDEs with a recurrent spatio-temporal network that encodes every agent’s actions for the duration of a simulation. This allows gradients to flow across both time and space from each spatio-temporal location to every action that affects it. As a result, training includes the context of all prior actions from across the physical space. We refer to this method as \emph{Backpropagation through Time and Space} (BPTTS). BPTTS can learn a generalizable solution by efficiently leveraging all available samples across the network.

Our application focus is similar to the work by \textcite{wang2019learning} which considers the problem of generating spatially higher-order accurate numerical schemes for hyperbolic PDEs. Observations are derived from the current state of the physics which we use to train a numerical policy to accurately advance the simulation. More specifically, we focus on learning higher order accurate flux reconstruction similar to state-of-the-art numerical schemes like Weighted Essentially Non-Oscillatory (WENO)~\parencite{Liu1994WeightedEN,shu1998essentially}. This approach contrasts with the current focus in scientific machine learning, which is to harness the representative power of neural networks \parencite{michoski-a,bar-sinai-a,r-a,weinan2017a,raissi2019a,li2020fourier} to approximate the solution, its gradients, or the operators in a PDE. The aim is also not to replace the entire numerical modeling pipeline with a neural network but instead to focus on the numerical algorithm, which is the most impactful, generalizable, and explainable piece.

With BPTTS, training converges to a policy that produces actions like SOTA methods. Learning in a large spatio-temporal network would appear expensive, but because the policy is applied on a local scale, it can readily be transitioned to situations other than where the learning was performed. For example, a policy that generalizes can be carried forward to study the physics in different embodiments by changing the physical extent, boundary, or initial conditions. 

For this work, training is guided by rewards based on error with the WENO scheme of \textcite{jiang1996weno}. This reward function is convenient for demonstrating our methods, but we recognize that this limits the accuracy of the solution predicted by the learned policy to at most that of the WENO scheme. As such, we emphasize that this work is not meant as an improvement upon WENO, but instead a demonstration of RL methods for this problem. With careful design of more general reward signals, better numerical methods could be learned for arbitrary conservation laws.

In this work, we introduce BPTTS, a novel method by which we train a spatio-temporal network that represents the breadth and duration of a physical simulation and contains homogeneous RL agents at every discretized location and time over the spatial volume. BPTTS resolves the non-stationarity that arises from multiple agents acting simultaneously for physical simulation. We show empirically that the policy learned by BPTTS predicts solutions that are comparable to SOTA methods and generalizes across grid discretizations and initial conditions.

\section{Related Work}

Homogeneous multi-agent reinforcement learning has been applied in scientific computing with an aim to either learn improved models \parencite{novati2021automating,bae2021scientific} or replace state-of-the-art numerics with learned policies \parencite{wang2019learning}. The experience from all agents is collected into a replay buffer \parencite{lin1992self} from which batches are drawn and a policy gradient is computed. This breaks the temporal correlation between experiences; this is preferred by stochastic gradient descent techniques as they rely on i.i.d. samples for convergence, but it loses valuable contextual information needed to account for local non-stationarity.
\textcite{wang2019learning}, to the best of our awareness, does not attempt to mitigate non-stationarity, and we find that their methods do not reliably learn behavior that is comparable to the SOTA solution.
\textcite{novati2021automating} mitigate non-stationarity with Ref-ER, or Remember and Forget Experience Replay \parencite{novati2019remember}, which can control the pace of policy changes. Restricting policy updates reduces the rate at which the perceived environment changes, allowing the policy to track the environment despite its non-stationarity. However, this approach is ad-hoc and requires tuning the hyperparameters that control the pace of learning. 

\textcite{mordatch2017emergence} train multiple homogeneous agents to communicate in a simple environment by encoding the episode as a single end-to-end network. We adapt this idea by extending the network to encompass agents at every spatial location for BPTTS. This resolves non-stationarity because training uses the information of all agents, an idea known as centralized training for decentralized execution \parencite{foerster2016learning}. Another approach to centralized training is to use some form of centralized value function, either a single unified function \parencite{rashid2018qmix,foerster2018counterfactual} or one for each agent \parencite{lowe2017multi}. Such an approach could be effective for this application, but we prefer the end-to-end network because it leverages the additional information available from the transition and reward functions known a priori.

Our approach trains the policy with a recurrent neural network (RNN). RNNs are an established technique in RL for handling partially observable environments \parencite{hausknecht2015deep}, including MARL environments \parencite{foerster2016learning}. They construct value and policy networks from LSTMs and GRUs, recurrent cells that have explicit mechanisms for maintaining memory across multiple timesteps. The RNN of our approach is distinct in that it has no explicit memory. Any information must be passed through the output of the cell, i.e., the physical simulation state in our application. This makes sense for the first-order differential equations under consideration, which depend only on the current state, and means the agent must act using information that is both spatially and temporally local. This allows a natural connection to existing numerical modeling techniques. If this requirement is relaxed or we wish to use higher orders, it may make sense to augment the RNN of our approach with explicit memory.

\section{Preliminaries}
\label{sec:background}

In this section, we provide a description of the simulation context in which learning is performed. Our interest is to evolve a conserved quantity, $u \in \mathbb{R}^d$, given over a uniform $N$-point discretization $D_N=\{x_1, x_2,\ldots,x_j,\ldots, x_N\}$ according to the hyperbolic PDE of the form: 

\begin{equation}
\frac{\partial u}{\partial t} + \frac{\partial}{\partial x}f(u) = 0
\end{equation}

where $f(u)$ are conserved fluxes that are exchanged at the interface, $x_{j\pm\half}$ of each cell, $I_j= [x_{j-\half},x_{j+\half}]$. The state $u(x_j,0)$ at each of the discrete points at the beginning of the simulation along with boundary values, $u(x_1, t)$ and $u(x_N, t)$ at all times are provided to complete the specification. The method of lines is then used to frame the PDE as a system of ordinary differential equations, where we approximate the spatial derivative with the finite difference:
\begin{equation}
\label{eqn:fd}
\frac{d u_j(t)}{d t} = - \frac{1}{\Delta x}(\fhat_{j+\half} - \fhat_{j-\half})
\end{equation}
$u_j$ is the approximation to the point value $u(x_j,t)$, and $\fhat_{j\pm\half}$ are numerical fluxes which need to be computed at the interfaces using cell values, possibly at higher spatial accuracy. With the approximated spatial derivatives, we can use a time-marching scheme like Euler's method to integrate in time. To compute the numerical fluxes, we use weighted essentially non-oscillatory (WENO) schemes. A WENO scheme of order $r$ achieves a $2r-1$ order spatially accurate construction of $\fhat_{j+\half}$ using the stencil $S_j$ of $2r-1$ points around $u(x_j,t)$. Specifically, it computes  $\fhat_{j+\half}$ as a convex combination of polynomials defined on $r$ sub-stencils inside $S_j$:
\begin{equation}
\fhat_{j+\half} =\sum_{k=0}^{r-1} \omega_k \fhat_{k,j+\half}
\label{eqn:weno_weights}
\end{equation}
where $\fhat_{k,j+\half}$ is the polynomial reconstruction of the $k$th sub-stencil.

A standard WENO scheme chooses the convex weights $\omega_k$ as:
\begin{equation}
\omega_{k}=\frac{\alpha_{k}}{\sum_{m=1}^{r} \alpha_{m}} \text { with } \alpha_{k}=\frac{d_{k}}{\left(\varepsilon+\beta_{k}\right)^{p}}
\label{eqn:weno_scheme}
\end{equation}
The weights are chosen to produce higher order approximations in smooth regions while reverting to low accuracy sub-stencil polynomials in regions of strong discontinuities to avoid spurious oscillations. $d_k$ is a pre-computed optimal coefficient which is adjusted by computing a smoothness indicator ($\beta_k$), and $\varepsilon$ and exponent $p$ are inputs. When all the smoothness indicators are the same magnitude, the weights will match the optimal weights, but when a smoothness indicator is large (such as at discontinuities), the contribution of its associated stencil will be small. Variants of the WENO scheme~\parencite{jiang1996weno, martin2006bandwidth, taylor2007optimization, arshed2013minimizing} use different choices for the sub-stencil weights $\omega_k$. The interest is in developing schemes that can sense the local solution state, for example where the derivatives of the solution vanish, and determine the weights to produce interface fluxes with the highest accuracy. This makes it a fruitful problem to study with machine learning techniques that derive mappings from the local state of physics to $\omega_k$.

\section{RL Formulation}
\label{sec:rl_formulation}
To apply reinforcement learning, we represent this setting as a partially observable Markov game (POMG) \parencite{littman1994markov}, a multi-agent extension of the Markov decision process. Plausibly, a single agent could act on the entire physical volume at once; however, this is impractical, as training a policy with hundreds of dimensions (or more) is difficult or impossible, and inflexible, as it constrains the input size to the grid discretization size $N$ used during training. Moreover, training a global agent ignores the inherent spatial independence of physics; the local PDE evolves the same irrespective of its location. Thus, we frame the problem as a POMG with many agents.

A POMG of $n$ agents has an internal state $s^t$. Observations $o_1^t,\ldots,o_n^t$ are computed from the internal state with observations functions $\mathcal{O}_i(s^t)=o_i$. For our setting, each observation function is deterministic. The agents select actions $a_1^t,\ldots,a_n^t$. A transition function $\mathcal{I}$ computes the following state, $s^{t+1}=\mathcal{I}(s^{t},a_1^t,\ldots,a_n^t)$, and a reward function $\mathcal{R}$ computes rewards for each agent, $r_1^t\ldots,r_n^t=\mathcal{R}(s^{t},a_1^t,\ldots,a_n^t)$. For our setting, each agent learns a deterministic policy $\pi$ mapping observations to actions, $\pi(o^t_i)=a^t_i$. In particular, we consider a partially observable Markov game that is homogeneous, that is, each agent shares parameters and uses an identical policy $\pi$. The objective is to find $\pi$ that maximizes the total return for all $n$ agents: 
\begin{equation}
	\max_{\pi} R(\pi) \quad \text{where} \quad R(\pi)=\sum_{t=0}^{T-1} \sum_{j=1}^{n} r_i^t
	\label{eqn:reward_sum}
\end{equation}

In our setting, the POMG has $N+1$ agents, one for each interface between cells in the discretized grid. The $j$th agent operates at the $j+\half$ interface. The internal state $s$ is given by the physical state $u(x,t)$ of the simulation. The $j$th agent observes numerical fluxes in stencil $S_j$ and chooses sub-stencil weights $\omega_k$ needed to advance the simulation to the next timestep. To ensure numerical stability and to avoid entropy-violating solutions, the flux is split into two parts, $f^+(u)$ and $f^{-}(u)$, using Lax-Friedrichs type flux splitting schemes~\parencite{lax1954weak}, where $f^\pm(u) =\half(f(u)\pm\alpha u)$ with $\alpha\geq \max |\partial_u f|$, then after reconstruction $\fhat_{j+\half}=\fhat^{+}_{j+\half}+\fhat^{-}_{j+\half}$. We focus on third order ($r=3$) WENO schemes; thus the agent observes $o_j^t=(f^+_{j-2},\ldots f^+_{j+2}),(f^-_{j-1},\ldots,f^-_{j+3})$ and learns a policy that provides two sets of weights $a_j^t=(\omega^+_1,\omega^+_2,\omega^+_3),(\omega^-_1,\omega^-_2,\omega^-_3)$ used in Equation~\ref{eqn:weno_weights}.

We consider episodes with finite length $T$ equal to the simulation time. The control interval between the agents' actions ($\Delta t$) is supplied by the CFL~\parencite{courant1967partial} constraint which states that for the simulation to be stable, the timestep must be less than the time it takes information to propagate across a single zone.

The reward is computed from the error with the standard WENO scheme of Equation~\ref{eqn:weno_scheme}. The reward at interface $j+\half$ is the average of the error in the two adjacent cells,
\begin{equation}
	r^t_j=-\frac{|u^t_j - w^t_j| + |u^t_{j+1} - w^t_{j+1}|}{2}
\end{equation}
where $u^t_j$ is part of the simulation state that makes up $s_t$, and $w^t_j$ is the corresponding state of the standard WENO scheme. Importantly, the reward is Markovian, that is, it depends on only the most recent state and action, not on states and actions from earlier timesteps. We achieve this by computing the state of the WENO scheme one timestep diverged from the POMG state, $w^{t+1}=\mathcal{I}(u^t, \mu(o_0)\ldots \mu(o_N))$, where $\mu(o_i)$ represents the equivalent policy of the standard WENO scheme. This is as opposed to computing a fixed $w_0,\ldots,w_T$ by evolving the standard WENO scheme from the initial condition. Notably, computing errors with a fixed solution transforms this approach into a form of supervised learning, while using a Markovian error retains dependence between actions and later samples, a distinct characteristic of RL.

However, BPTTS accounts for the influence of past states and actions, so it is not obvious that Markovian rewards (and RL) are required. Despite this, we believe they are still important, based on our experiments that minimized error with a fixed solution, the results of which were poor and shown in Appendix~\ref{appendix:burgers_plot:supervised}. We suspect this is a consequence of chaotic temporal dependencies. The final state depends on the initial state in a way that is unpredictable and changes rapidly. A non-Markovian reward that depends on the initial state will also be chaotic, unlike a Markovian reward that only depends on the most recent state.

\section{Backpropagation through Time and Space}
\label{sec:methods}
Training multiple agents requires a careful approach. If we train the agents on their local observations and actions using policy gradients or Q-learning, the environment will appear non-stationary from the perspective of the individual agent. Consider this contrived example: the observation of agent $j_1$ at timestep $t_1$ happens to be the same as agent $j_2$ at future timestep $t_2$, $o_{j_1}^{t_1}=o_{j_2}^{t_2}$, so the policy chooses the same action $a$ at both points. Based on experience, the policy adjusts to choose action $b$ instead. In this example, the two observations stay the same, so both agents switch to $b$. Agent $j_1$ receives the expected increase in reward, but the effect of this change in action propagates over time and space until it affects the reward for agent $j_2$ at $t_2$ but not the observation, substantially decreasing the reward instead. From the unified agent's perspective, this makes no sense: for two identical observations, choosing $b$ in one case increases the reward but in the other case decreases it. In the worst case, this experience might update the policy to choose $a$ again.

\begin{wrapfigure}{R}{0.5\linewidth}
	\centering
	\includegraphics[width=1.0\linewidth]{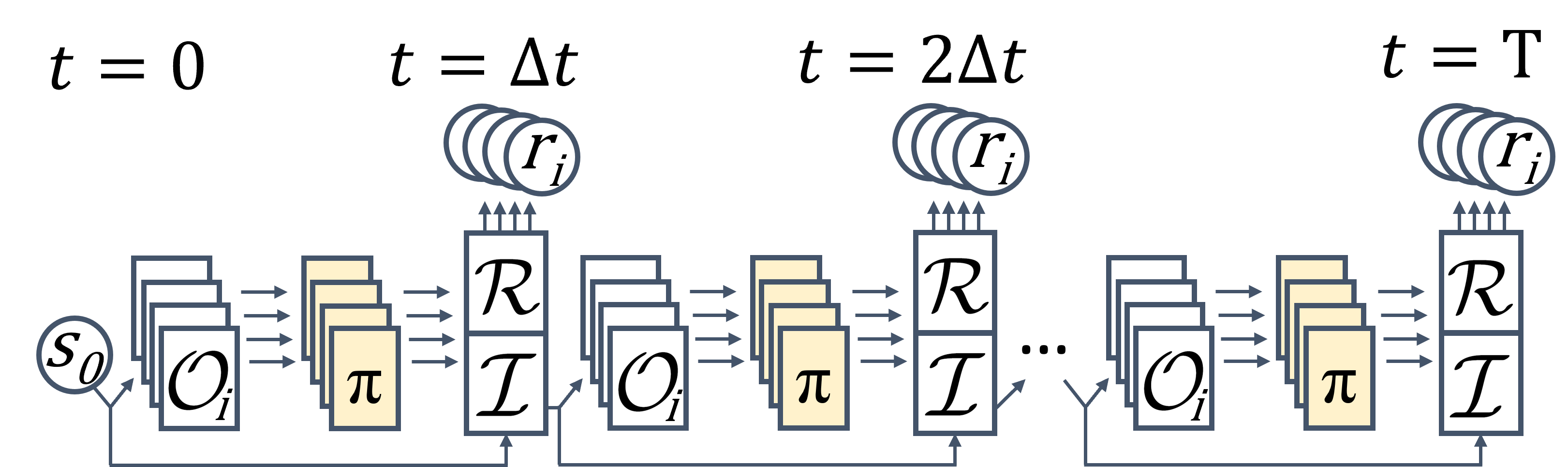}
	\caption{The recurrent spatio-temporal network trained with BPTTS. $s_0$ is the initial condition of the PDE, the observation functions $\mathcal{O}_i$ compute the flux in each stencil of the state, each instance of the policy $\pi$ choose an action of sub-stencil weights, the transition function $\mathcal{I}$ uses the sub-stencil weights and the previous state to integrate forward to the next state, and the reward function $\mathcal{R}$ computes rewards for each agent based on the error in the next state. The network is trained to maximize the rewards. Only the yellow boxes representing the policy contain trainable parameters.}
	\label{fig:global_backprop}
\end{wrapfigure}

Such an approach fails because the isolated experience of each agent is insufficient for learning. Learning must be centralized and include the context of the actions and observations of the other agents, even while training a decentralized agent that needs only local information to act. One way to do this is to encode the entire duration of an episode into a neural network. This is possible because, unlike RL environments in general, we know the transition and reward functions of the PDE environment a priori.
The network is recurrent and made up of many identical cells laid end-to-end. Each cell contains the observation functions, each instance of the policy, and the transition and reward functions. This is pictured in Figure~\ref{fig:global_backprop}. We refer to training this network as \emph{Backpropagation through Time and Space} (BPTTS) because gradients flow from each reward to the actions at each time and location that influences that reward. The policy trained with BPTTS is still a function of only the local observation, but trains with the benefit of globally accumulated gradients.

The input of the BPTTS network is the initial state, or the initial condition of the equation, and the output is the rewards and states at each timestep. We train the parameters $\theta$ of the network to maximize the sum of all rewards across space and time, that is, the sum of returns $R(\pi)$ in Equation~\ref{eqn:reward_sum}. Gradients are accumulated for each instance of the policy and combined for the total gradient:
\begin{equation}
	\nabla_\theta R(\pi|\theta)=
	\sum_{s,j}\left(\nabla_{\theta_{s,j}}\sum_{t,i}r_i^t\right)
\end{equation}
where $\nabla_{\theta_{s,j}}$ refers to the gradients with respect to the parameters of the policy at time $s$ and grid location $j$. For each $(s,j)$ there are many $r_i^t$ that are unaffected by that action and for which the gradient will be $0$, such as when $t<s$. Maximizing $R(\pi)$ is analogous to maximizing the full undiscounted return, $\sum_{\tau=t}^T r_\tau$, used as an objective for policy gradients \parencite{sutton2018reinforcement}, except $R(\pi)$ accounts for rewards across space as well as time. 

\section{Experiments}
BPTTS is applicable to any hyperbolic conservation law. In this section we show results on the Burgers' equation and the Euler equations. We are interested in how the policy improves over the course of training and how well the trained policy performs relative to the standard WENO scheme. As the standard WENO scheme is used for the reward signal, we consider matching the behavior of the WENO scheme as a successful outcome, even if WENO diverges somewhat from the true solution. We also evaluate how the policy generalizes to different spatial discretizations and to different initial conditions.

Additional experimental details are available in the appendices. Hyperparameters and other training details are in Appendix~\ref{appendix:training_details}. Additional experiments and plots are shown in Appendices~\ref{appendix:burgers_plot}~and~\ref{appendix:euler_plot} for the Burgers' equation and Euler equations respectively.

\subsection{Burgers' Equation}

\begin{wrapfigure}{R}{0.5\linewidth}
	\includegraphics[width=1.0\linewidth]{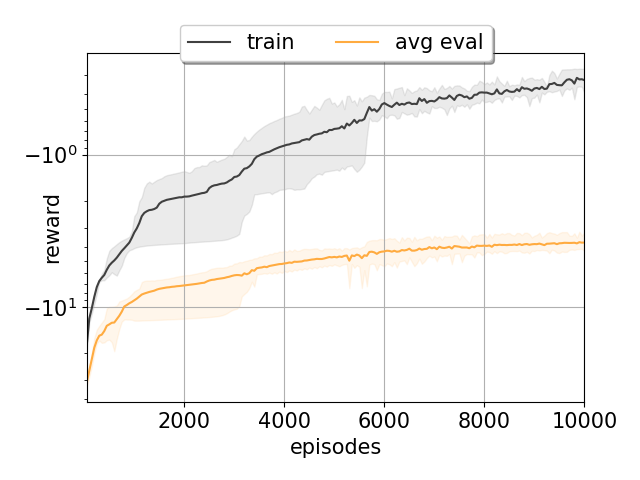}
	\vspace{-0.2in}
	\caption{Reward while training on the Inviscid Burgers Equation.  This plot is an average of 6 runs with different random seeds. The shaded region represents the highest and lowest value between the seeds at that episode.}
	\label{fig:burgers:training}
\end{wrapfigure}

In this section, we demonstrate BPTTS on the inviscid Burgers' equation, a model non-linear hyperbolic equation. It is given by the following PDE:
\begin{equation}
	\frac{\partial u}{\partial t} + \frac{1}{2}\frac{\partial u^2}{\partial x} = 0
\end{equation}
where flux is $f(u) = \frac{1}{2}u^2$. We show that BPTTS learns to simulate the Burgers' equation similar to the standard WENO solution and can generalize to unseen initial conditions and grid sizes.

We train the BPTTS network with batches of three initial conditions. Each batch includes the same three initial conditions pictured in Figure~\ref{fig:burgers:eval:seen}. These initial conditions are chosen because they contain between them a variety of types of physics with which to generalize to other initial conditions.

During training, each initial condition is evolved for 250 timesteps of 0.0004 seconds, for a total duration of 0.1 second; the recurrent network has 250 repeated cells. During evaluation, we evolve to 500 timesteps for a total of 0.2 seconds to show that the learned policy predicts the long-term evolution of physics. We also evaluate the policy on other initial conditions not seen during training (Figure~\ref{fig:burgers:eval:unseen}). During training, the space is discretized into $N=128$ cells. The policy is trained for 10,000 episodes of each training initial condition. The policy is evaluated every 50 episodes, and the instance of the policy with the highest total reward across the training environments is selected for further evaluation.

From Figure~\ref{fig:burgers:training}, we can see that the total reward increases over the course of training for the training environments, and it saturates for the evaluation environments. The total reward here is the sum of rewards across time and space. The maximum possible reward is $0.0$, which represents acting identically to the standard WENO solution. Note that the total training reward is equivalent to the objective function used to optimize the network. The reward on evaluation environments is expected to be somewhat less than for training environments, both from small subtleties in the unseen physics that the policy struggles to generalize to, and we run the evaluation environments for twice the duration, so they accumulate more negative rewards.

\begin{figure}[p]
	\centering
	\begin{subfigure}{0.32\linewidth}
		\includegraphics[width=\linewidth]{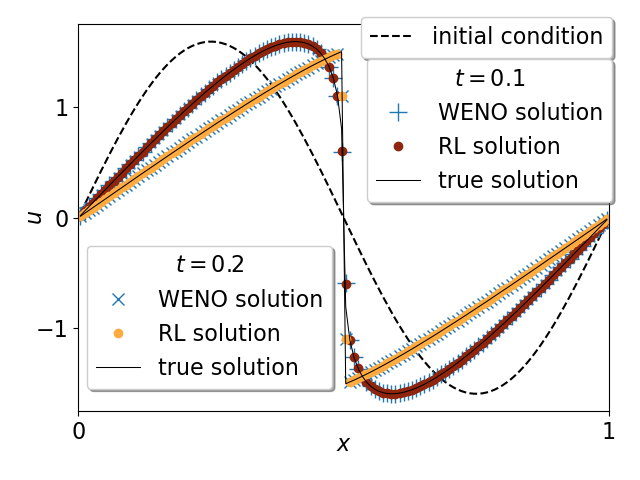}
		\vspace{-7mm}
		\caption{\textit{standing sine}}
		\label{fig:burgers:eval:sine}
	\end{subfigure}
	\begin{subfigure}{0.32\linewidth}
		\includegraphics[width=\linewidth]{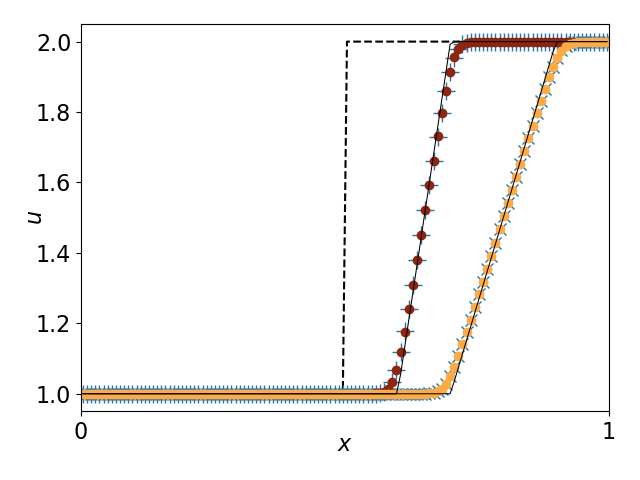}
		\vspace{-7mm}
		\caption{\textit{rarefaction}}
		\label{fig:burgers:eval:rarefaction}
	\end{subfigure}
	\begin{subfigure}{0.32\linewidth}
		\includegraphics[width=\linewidth]{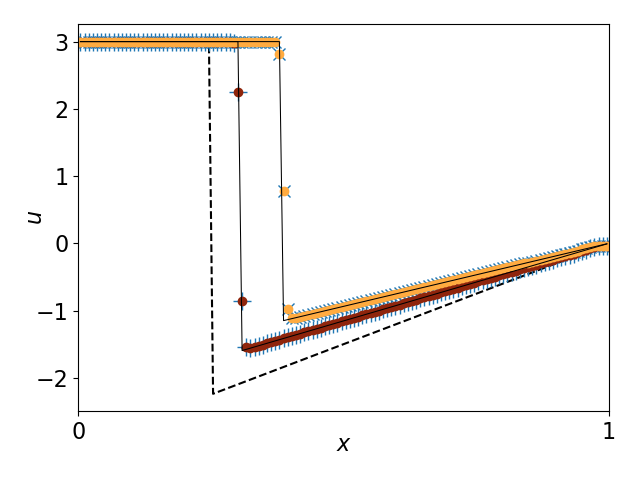}
		\vspace{-7mm}
		\caption{\textit{accelerating shock}}
		\label{fig:burgers:eval:accelshock}
	\end{subfigure}
	\customlabel{fig:burgers:eval:seen}{\ref{fig:burgers:eval}\subref{fig:burgers:eval:sine}-\subref{fig:burgers:eval:accelshock}}
	\newline
	\begin{subfigure}{0.32\linewidth}
		\includegraphics[width=\linewidth]{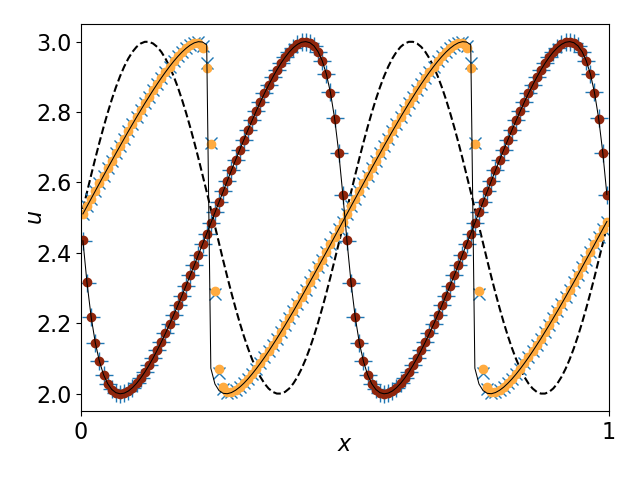}
		\vspace{-7mm}
		\caption{\textit{double sine}}
		\label{fig:burgers:eval:other}
	\end{subfigure}
	\begin{subfigure}{0.32\linewidth}
		\includegraphics[width=\linewidth]{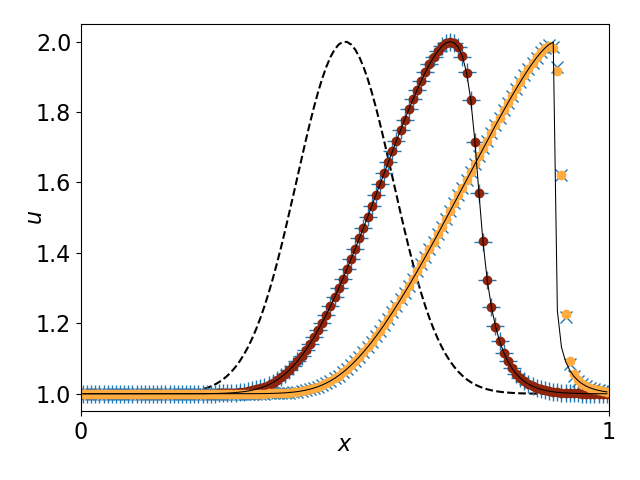}
		\vspace{-7mm}
		\caption{\textit{gaussian}}
		\label{fig:burgers:eval:gaussian}
	\end{subfigure}
	\begin{subfigure}{0.32\linewidth}
		\includegraphics[width=\linewidth]{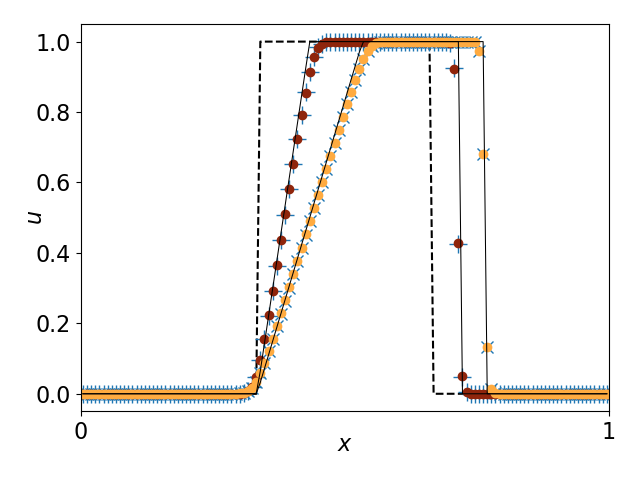}
		\vspace{-7mm}
		\caption{\textit{tophat}}
		\label{fig:burgers:eval:tophat}
	\end{subfigure}
	\customlabel{fig:burgers:eval:unseen}{\ref{fig:burgers:eval}\subref{fig:burgers:eval:other}-\subref{fig:burgers:eval:tophat}}

	\caption{Evaluation of the policy trained on the Inviscid Burgers' Equation. Initial conditions are evolved to 0.2s, twice the duration of a training episode. (\subref{fig:burgers:eval:sine}-\subref{fig:burgers:eval:accelshock}) are used during training, (\subref{fig:burgers:eval:other}-\subref{fig:burgers:eval:tophat}) are not seen by the agents until evaluation.}
	\label{fig:burgers:eval}
\end{figure}

\begin{figure}[p]
	\centering
	\includegraphics[width=\linewidth]{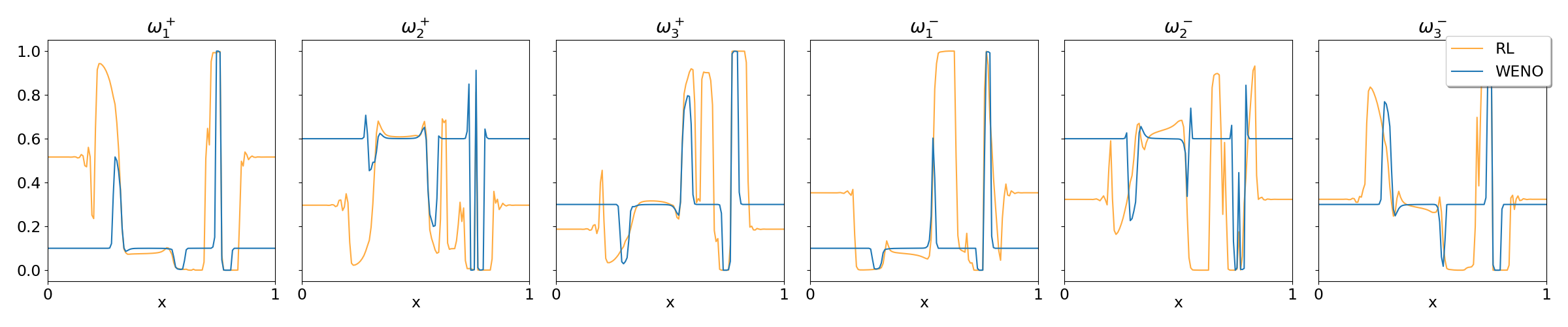}
	\vspace{-7mm}
	\caption{Comparison of actions (that is, the sub-stencil weights) of the policy and of the standard WENO scheme. These actions are from the second-to-last timestep of the \textit{tophat} initial condition in Figure~\ref{fig:burgers:eval:tophat}. (\textit{tophat} is chosen as it best highlights that the learned weights are different.)}
	\label{fig:burgers:actions}
\end{figure}

\begin{table}[p]
	\centering
	\caption{Comparison of the L2 error with the true solution on the Burgers' equation for the RL policy and the standard WENO scheme.}
	\vspace{7.5pt}
	\resizebox{1.0\linewidth}{!}{
		\begin{tabular}{|l|ll|ll|ll|ll|ll|ll|}
			\hline
			& \multicolumn{2}{l|}{\textit{standing sine}}       & \multicolumn{2}{l|}{\textit{rarefaction}}         & \multicolumn{2}{l|}{\textit{accelerating shock}} & \multicolumn{2}{l|}{\textit{double sine}}         & \multicolumn{2}{l|}{\textit{gaussian}}            & \multicolumn{2}{l|}{\textit{tophat}} \\ \hline
			$N$               & \multicolumn{1}{l|}{\textbf{RL}}     & \textbf{WENO}       & \multicolumn{1}{l|}{\textbf{RL}}     & \textbf{WENO}       & \multicolumn{1}{l|}{\textbf{RL}}      & \textbf{WENO}       & \multicolumn{1}{l|}{\textbf{RL}}     & \textbf{WENO}       & \multicolumn{1}{l|}{\textbf{RL}}     & \textbf{WENO}       & \multicolumn{1}{l|}{\textbf{RL}}     & \textbf{WENO}       \\ \hline
			64                & \multicolumn{1}{l|}{0.0719} &  0.0712 & \multicolumn{1}{l|}{0.0207} & 0.0209 & \multicolumn{1}{l|}{0.1856}  & 0.1853 & \multicolumn{1}{l|}{0.1149} & 0.1137 & \multicolumn{1}{l|}{0.0500} & 0.0476 & \multicolumn{1}{l|}{0.0584} & 0.0581 \\ \hline
			128               & \multicolumn{1}{l|}{0.0506} & 0.0504 & \multicolumn{1}{l|}{0.0114} & 0.0115 & \multicolumn{1}{l|}{0.1723}  & 0.1721 & \multicolumn{1}{l|}{0.0869} & 0.0867 & \multicolumn{1}{l|}{0.0750} & 0.0756 & \multicolumn{1}{l|}{0.0326} & 0.0325 \\ \hline
			256               & \multicolumn{1}{l|}{0.0357} & 0.0356 & \multicolumn{1}{l|}{0.0062} & 0.0062 & \multicolumn{1}{l|}{0.0684}  & 0.0683 & \multicolumn{1}{l|}{0.0953} & 0.0961 & \multicolumn{1}{l|}{0.0690} & 0.0695 & \multicolumn{1}{l|}{0.0357} & 0.0357 \\ \hline
			512               & \multicolumn{1}{l|}{0.0252} & 0.0252 & \multicolumn{1}{l|}{0.0033} & 0.0033 & \multicolumn{1}{l|}{0.0546}  & 0.0545 & \multicolumn{1}{l|}{0.1144} & 0.1149 & \multicolumn{1}{l|}{0.0737} & 0.0739 & \multicolumn{1}{l|}{0.0315} & 0.0316 \\ \hline
			1024              & \multicolumn{1}{l|}{0.0178} & 0.0178 & \multicolumn{1}{l|}{0.0017} & 0.0017 & \multicolumn{1}{l|}{0.0503}  & 0.0503 & \multicolumn{1}{l|}{0.1136} & 0.1138 & \multicolumn{1}{l|}{0.0796} & 0.0797 & \multicolumn{1}{l|}{0.0137} & 0.0137 \\ \hline
		\end{tabular}
		\label{table:burgers_test}
	}
	\vspace{-3mm}
\end{table}

\begin{figure}[t]
	\centering
	\begin{minipage}{0.50\linewidth}
		\centering
		\begin{subfigure}{0.54\linewidth}
			\includegraphics[width=\linewidth]{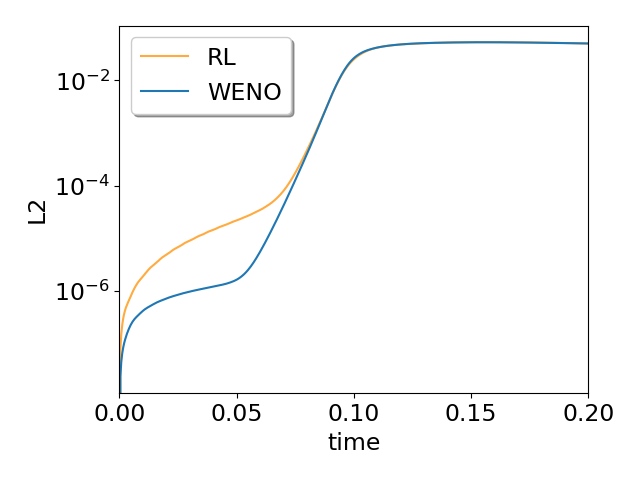}
			\caption{}
			\label{fig:burgers:l2}
		\end{subfigure}
		\begin{subfigure}{0.44\linewidth}
			\includegraphics[width=\linewidth]{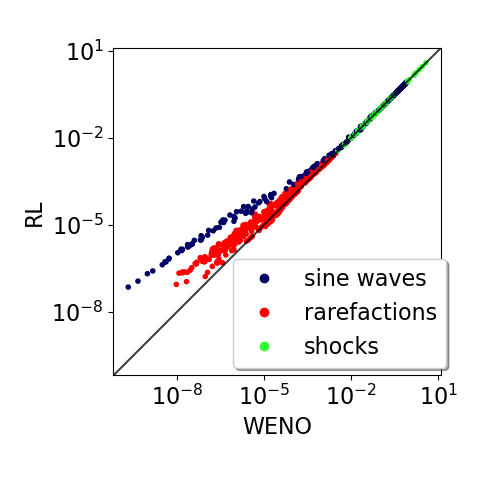}
			\vspace{-7mm}
			\caption{}
			\label{fig:burgers:error}
		\end{subfigure}
		\vspace{0mm}
		\caption{Comparison of the L2 error with the true solution for the trained policy and the standard WENO scheme. (\subref{fig:burgers:l2}) shows the error over the course of an episode starting from the standing sine initial condition (Figure~\ref{fig:burgers:eval:sine}). (\subref{fig:burgers:error}) compares error across many random environments with random grid sizes. Both show the policy's focus on high error conditions, where it matches the WENO scheme closely.}
	\end{minipage}
	\hfill
	\begin{minipage}{0.45\linewidth}
		\centering
		\includegraphics[width=1.0\linewidth]{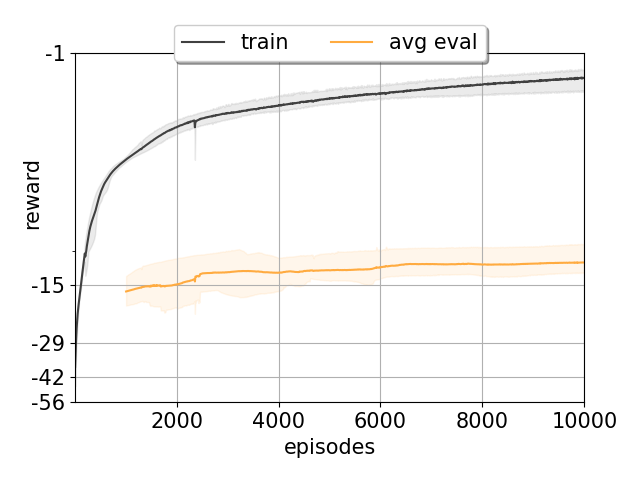}  
		\vspace{-8mm}
		\caption{Reward while training on Sod initial condition for Euler Equations. This plot is an average of 6 runs with different random seeds.}
		\label{fig:euler_reward}
	\end{minipage}
\end{figure}

The learned policy learns to closely mimic the solution of the standard WENO scheme, see Figure~\ref{fig:burgers:eval}. Where the WENO scheme diverges from the true solution the policy also diverges. Notably, the policy matches the solution of the WENO scheme but does not use the same actions; Figure~\ref{fig:burgers:actions} shows that the sub-stencil weights selected by the policy are similar but different to those of the WENO scheme. Instead of simply cloning the behavior of the WENO scheme, the policy learns about the physics and chooses weights to minimize error. As the plots of Figure~\ref{fig:burgers:eval} are only two instants in time, we also include Figure~\ref{fig:burgers:l2} to show the performance of the learned policy throughout the episode.
From this, we can see that the policy prioritizes states where the error of the standard WENO scheme is relatively high, matching WENO closely; in states where the WENO scheme has lower error, the policy has higher but acceptable errors.

Besides sample efficiency, an advantage of treating the problem as MARL with many local agents is that we can use the learned policy with different grid discretizations by simply continuing to apply the policy at every grid interface. We compare the error with the true solution for the learned policy and the WENO scheme, across several grid sizes for Table~\ref{table:burgers_test}, and for 1,200 randomized environments in Figure~\ref{fig:burgers:error}. Table~\ref{table:burgers_test} shows that the policy generalizes well to different grid sizes for the evaluation environments. For Figure~\ref{fig:burgers:error}, each environment samples $N$ from a discrete log uniform distribution between $64$ and $1024$, then samples the numerical constants of the initial condition (details are in Appendix~\ref{appendix:train:init}).
From this plot, we see a similar pattern to Figure~\ref{fig:burgers:l2}: the policy matches the standard WENO scheme when error is high and has somewhat higher errors than the WENO scheme when error is low.

\subsection{Euler Equations}
In this section, we consider the Euler equations, which describe the conservation of mass, momentum, and energy for fluids. They are given by:

\begin{equation}
	\mathcal{U}_{t}+[\mathbf{F}(\mathcal{U})]_{x}=0 
\end{equation}
with
\begin{equation}
	\begin{gathered}
		\mathcal{U}=\left(\begin{array}{c}
			\rho \\
			\rho u \\
			\rho E
		\end{array}\right) \quad \mathbf{F}(\mathcal{U})=\left(\begin{array}{c}
			\rho u \\
			\rho u u+p \\
			\rho u E+u p
		\end{array}\right)
	\end{gathered}
\end{equation}
where $\rho$ is the density, $u$ is the velocity, $p$ is the pressure, and $E$ is the total energy. $E$ is expressed in terms of the specific internal energy $e$ and kinetic energy as:
\begin{equation}
	E = e + \frac{1}{2}u^2
\end{equation}
The equations are closed by the addition of an equation of state; for this environment we use the gamma-law given by $p=\rho e (\gamma-1)$ where $\gamma=1.4$.

We use BPTTS for the Euler equations similar to how it is used for the Burgers' equation. Each of the three equations is handled separately. The agents have the same sized observation and action spaces as with the Burgers' equation, except there are three homogeneous agents at every spatio-temporal location instead of one. With this adjustment, gradients may flow not just across time and space but also between different equations. The reward is averaged over the three equations.

\begin{figure}[t]
	\centering
	\begin{subfigure}{0.32\linewidth}
		\includegraphics[width=\linewidth]{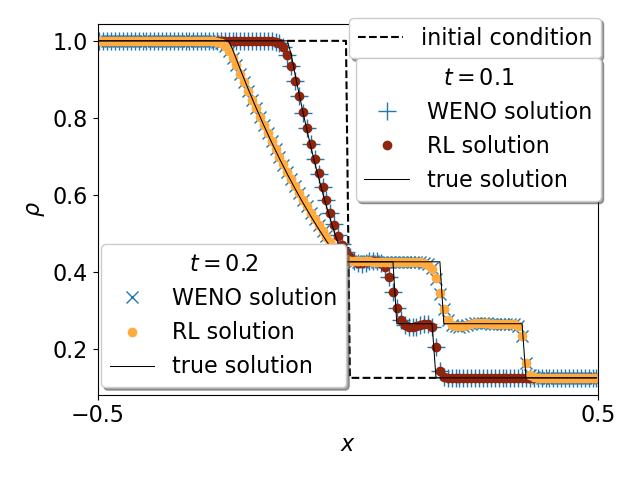}
	\end{subfigure}
	\begin{subfigure}{0.32\linewidth}
		\includegraphics[width=\linewidth]{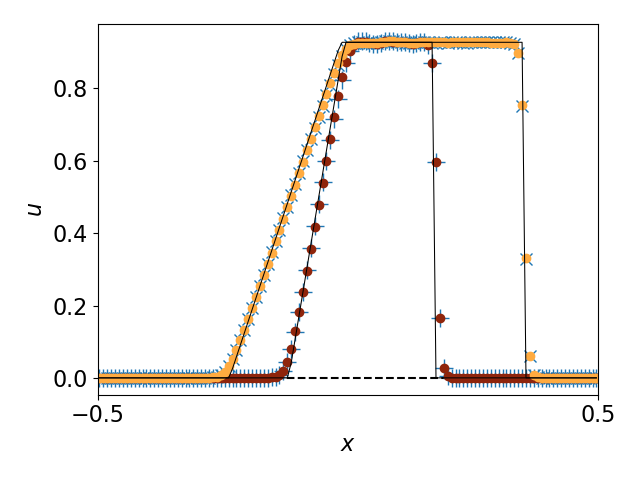}
	\end{subfigure}
	\begin{subfigure}{0.32\linewidth}
		\includegraphics[width=\linewidth]{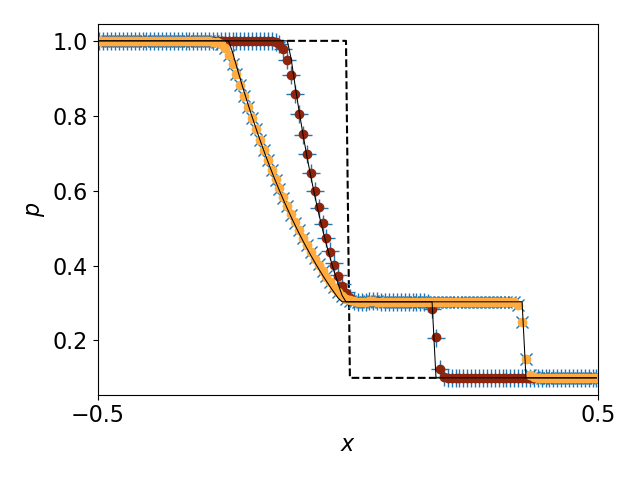}
	\end{subfigure}
	\vspace{-5mm}
	\caption{Evaluation of the trained policy on the Sod problem. It is evolved to 0.1s during training, and 0.2s during evaluation.}
	\label{fig:euler:sod_state}
\end{figure}

As the space of possible initial conditions for the Euler equations is broad, we constrain our study to shock tubes with a single discontinuity separating two states. The initial discontinuity resolves into three waves whose magnitude and propagation speed depend on the initial values of the left and right states.
Unlike with the Burgers' equation, each training step has only one initial condition: the Sod problem~\parencite{sod1978survey}, pictured in Figure~\ref{fig:euler:sod_state}. The values for the left and right states for the Sod problem and other initial conditions are given in Appendix~\ref{appendix:training_details}. The Sod problem itself contains sufficiently varied physics that allow the learned policy to generalize among initial conditions.

During training, the Sod problem is evolved to 1,000 timesteps of 0.0001 seconds for a total of 0.1 seconds. We evaluate the learned policy on the Sod problem evaluated to 0.2 seconds, as well as three additional initial conditions: the Sod2, Lax, and Sonic Rarefaction problems~\parencite{masatsuka2013like}. As with the Burgers' equation, we use $N=128$ cells, train for 10,000 episodes, and evaluate every 50 episodes.

Figure~\ref{fig:euler_reward} shows the total reward over the course of training on the Sod problem. The Sonic Rarefaction problem is sensitive to errors and explodes with an untrained policy, so we only show the average evaluation reward after the first 1,000 episodes. As with the Burgers' equation, the reward on the training environment increases over the course of learning, while the reward for the evaluation environments saturates.

We analyze the ability of the learned policy to generalize to different grid discretizations by comparing it to the standard WENO scheme on several different grid sizes $N$. These results are presented in Table~\ref{table:euler_test}. The L2 error with the true solution is similar between the learned policy and the WENO scheme, showing that the policy generalizes to different grids for the Euler equations.

The learned policy closely follows the comparison WENO scheme. For the Sod problem, this is shown in Figure~\ref{fig:euler:sod_state}; the remaining initial conditions as well as action plots similar to Figure~\ref{fig:burgers:actions} are presented in Appendix~\ref{appendix:euler_plot}.

\begin{table}[t]
	\centering
	\caption{Comparison of the L2 error with the true solution on the Euler equations for the RL policy and the standard WENO scheme.}
	\vspace{7.5pt}
	\resizebox{0.7\linewidth}{!}{
		\begin{tabular}{|l|l|l|l|l|l|l|l|l|}
			\hline
			& \multicolumn{2}{l|}{\textit{Sod}} & \multicolumn{2}{l|}{\textit{Sod2}} & \multicolumn{2}{l|}{\textit{Lax}} & \multicolumn{2}{l|}{\textit{Sonic Rarefaction}} \\
			\hline
			$N$ & \textbf{RL} & \textbf{WENO} & \textbf{RL} & \textbf{WENO}& \textbf{RL} & \textbf{WENO}& \textbf{RL} & \textbf{WENO}\\
			\hline
			64  & 0.0575  & 0.0576 & 0.0454 & 0.0463 & 0.4140 & 0.3930 & 1.4254 & 1.3922\\ 
			\hline
			128 & 0.0318 & 0.0321 & 0.0289 & 0.0292 & 0.3571 & 0.3527 & 0.9251 & 0.9046 \\
			\hline
			256 & 0.0221 & 0.0221 & 0.0189 & 0.0191 & 0.1935 & 0.1873 & 0.5703 & 0.5553 \\
			\hline
			512 & 0.0181 & 0.0179 & 0.0131 & 0.0132 & 0.1930 & 0.1910 & 0.4454 & 0.3986 \\
			\hline
		\end{tabular}
		\vspace{-5mm}
		\label{table:euler_test}
	}
\end{table}

\section{Conclusion}
In this work, we have demonstrated Backpropagation Through Time and Space, a technique for learning numerical methods for hyperbolic conservation laws with homogeneous MARL. BPTTS solves the challenge of non-stationarity that arises from multiple agents acting simultaneously. We have further shown that a policy trained with BPTTS generalizes to different discretizations of the physical space and to different initial conditions of the PDEs.

This work opens the possibility for future improvements to existing numerical methods. While the accuracy of our learned solution is limited by our use of the standard WENO solution for rewards, BPTTS is not restricted to this particular reward function, and we have shown that the policy learns an understanding of the physics, as opposed to simply cloning the standard WENO solution. With careful design of general intrinsic reward signals, BPTTS may learn numerical methods that outperform existing numerical methods on arbitrary conservation equations.

\section*{Acknowledgments}
This material is based upon work supported by the Defense Advanced Research Projects Agency (DARPA) under Agreement No. HR00112090063. Approved for public release; distribution is unlimited.

\printbibliography

\clearpage
\appendix
\onecolumn

\section{Training Details}
\label{appendix:training_details}
\subsection{Hyperparameters}
For both Euler equations and Burgers' equation experiments, the policy is encoded as a neural network with two hidden layers, each with 128 neurons. Both layers use the rectified linear unit (ReLU) for the activation function. Each network weight a 64-bit floating point number. We use the Adam algorithm \parencite{kingma2014adam} to optimize the network, with a learning rate of $0.0003$.

\subsection{Burgers' Equation Initial Conditions}
\label{appendix:train:init}

The details of the initial conditions used in the Burgers' equation are listed in Table~\ref{table:burgers_init}. The boundary refers to how the boundary conditions are handled for generating observations near the edge of the grid. For `periodic,' additional cells are copied from the opposite side of the grid. For `outflow,' the outermost cells are repeated as necessary.

The bottom three rows of Table~\ref{table:burgers_init} are used to generate Figure~\ref{fig:burgers:error}. 400 of each type are generated. The constants $a,b,c,k,\phi$ are sampled from uniform distributions to vary the initial conditions. Specifically, for random sines $a\sim[-1.0,-0.2]\cup[0.2,1.0],k\sim\{2,4,6,8,10\}$, for random shocks $c\sim[0.5,5],a\sim[0,5],\phi\sim[0,0.5]$, and for random rarefactions $c\sim[-1,1],a\sim[0.25,1.5],b\sim[20,100]$.

\renewcommand{\heavyrulewidth}{1.5pt}

\begin{table}[h!]
	\centering
	\caption{Initial conditions used during Burgers' equation training and evaluation.}
	\vspace{7.5pt}
	\begin{tabular}{p{0.22\linewidth}|l|c|c}
		Name & $u(x)=$ & Boundary & Fig. \\
		\toprule
		standing sine & $0.1 \frac{1}{2\pi}\sin(2\pi x)$ & periodic &
		\ref{fig:burgers:eval:sine} \\
		\midrule
		rarefaction &
		$\begin{cases} 1 & x\leq0.5 \\ 2 & x>0.5 \\	\end{cases}$ &
		outflow & \ref{fig:burgers:eval:rarefaction} \\
		\midrule
		accelerating shock &
		$\begin{cases} 3 & x\leq0.25 \\	3(x-1) & x>0.25 \\ \end{cases}$ &
		outflow & \ref{fig:burgers:eval:accelshock} \\
		
		\toprule
		double sine & $0.25+0.5\sin(4\pi x)$ & periodic &
		\ref{fig:burgers:eval:other} \\
		\midrule
		gaussian & $1+e^{-60(x-0.5)^2}$ &
		outflow & \ref{fig:burgers:eval:gaussian} \\
		\midrule
		tophat &
		$\begin{cases} 0 & x\leq 1/3 \\ 1 & 1/3<x<2/3 \\
			0 & x\geq 2/3 \\ \end{cases}$ &
		outflow & \ref{fig:burgers:eval:tophat} \\
		
		\toprule
		random sine & $3.5-|a|+a\sin(k\pi x)$ & periodic & \\
		\midrule
		random shock &
		$\begin{cases} c & x\leq\phi \\	a(x-1) & x>\phi \\ \end{cases}$ &
		outflow & \\
		\midrule
		random \linebreak rarefaction & $c+a\tanh(b(x-0.5))$ & outflow & \\
		
		\bottomrule
	\end{tabular}
	\label{table:burgers_init}
\end{table}

\subsection{Euler Equations Initial Conditions}
Each initial condition used for the Euler Equations is a shock tube with a single discontinuity separating two states. The initial left and right states and other parameters of each initial condition are given in Table~\ref{table:euler_init}.
$\rho_l$, $u_l$, and $p_l$ refer to the initial density, velocity, and pressure of the fluid on the left; $\rho_r$, $u_r$, and $p_r$ refer to the same on the right.

\begin{table}[h!]
	\centering
	\caption{Initial conditions used during Euler Equations training and evaluation}
	\vspace{7.5pt}
	\begin{tabular}{l|c|c|c|c|c|c}
		Name & Left  & Right & Boundary & $x_{min}$ & $x_{max}$ & $t_{max}$   \\
		\toprule
		Sod & 
		$\begin{array}{c}\rho_l=1 \\u_l=0 \\p_l=1\end{array}$ & 
		$\begin{array}{c}\rho_r=1/8  \\u_r=0 \\p_r=1/10\end{array}$ &
		outflow & $-0.5$ & $0.5$ & $0.2s$ \\
		\midrule
		Sod2 & 
		$\begin{array}{c}\rho_l=1 \\u_l=0 \\p_l=1\end{array}$ & 
		$\begin{array}{c}\rho_r=0.01  \\u_r=0 \\p_r=0.01\end{array}$ &
		outflow & $-0.5$ & $0.5$ & $0.2s$ \\
		\midrule
		Lax & 
		$\begin{array}{c}\rho_l=0.445 \\u_l=0.689 \\p_l=3.528\end{array}$ & 
		$\begin{array}{c}\rho_r=0.5  \\u_r=0 \\p_r=0.571\end{array}$ &
		outflow & $-0.5$ & $0.5$ & $0.14s$ \\
		\midrule
		Sonic Rarefaction & 
		$\begin{array}{c}\rho_l=3.857 \\u_l=0.92 \\p_l=10.333\end{array}$ & 
		$\begin{array}{c}\rho_r=1  \\u_r=3.55 \\p_r=1\end{array}$ &
		outflow & $-5.0$ & $5.0$ & $0.7s$ \\
		\bottomrule
	\end{tabular}
	\label{table:euler_init}
\end{table}

\section{Additional Burgers' Equation Experiments}
\label{appendix:burgers_plot}

\subsection{Fixed Solution Rewards}
\label{appendix:burgers_plot:supervised}

\begin{wrapfigure}{R}{0.5\linewidth}
	\centering
	\includegraphics[width=1.0\linewidth]{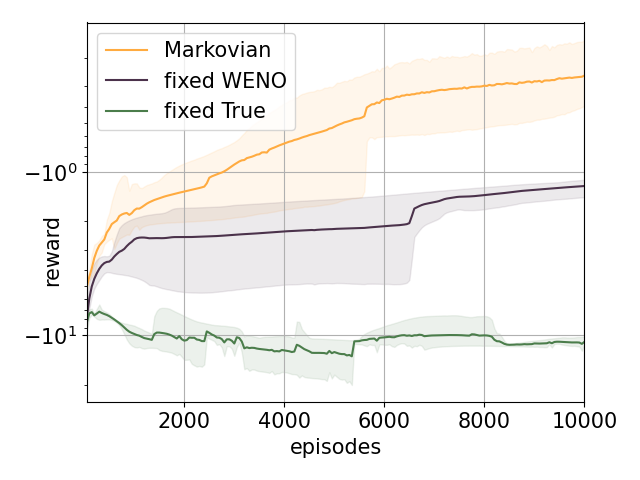}
	\caption{Training with a fixed solution. `Markovian' refers to training with the original unfixed reward function. `fixed WENO' refers to using the standard WENO scheme to compute the fixed solution, and `fixed True' refers to using true solution as the fixed solution. As these configurations change the reward function itself, the reward plotted here uses the original Markovian reward. Both configurations of fixed solutions are averages of 3 seeds.}
	\label{fig:burgers:supervised}
\end{wrapfigure}

In Section~\ref{sec:rl_formulation}, we describe the form of the reward function. The reward for an interface is computed as the error in adjacent cells,
\[r^t_j=-\frac{|u^t_j - w^t_j| + |u^t_{j+1} - w^t_{j+1}|}{2}\]
where $u^t_j$ is part of the physical state that makes up the POMG state $s_t$, and $w^t_j$ is part of the corresponding state of another solution. For our main results, the other solution is the standard WENO scheme one timestep diverged from the POMG state, that is, $w^{t+1}=\mathcal{I}(u^t, \mu(o_0)\ldots \mu(o_N))$, where $\mu(o_i)$ represents the equivalent policy of the standard WENO scheme. This reward is Markovian because it depends on the most recent state, not on earlier states. This is in contrast to a fixed solution $w_0,\ldots,w_T$ that stays the same for every episode. Rewards from a fixed solution are non-Markovian because they do not depend on the variable POMG state. However, as the network trained by BPTTS includes temporal connections, it is plausible that a non-Markovian reward will be effective, and we experimented with computing the reward from a fixed solution. We considered two such solutions: the standard WENO solution, except evolved from the initial condition instead of the previous POMG state, and the true solution.

Unlike the training for our main results, we found that training with the \textit{accelerating shock} initial condition (Figure~\ref{fig:burgers:eval:accelshock}) when using fixed solutions was impossible.
The long-ranged dependence between the final reward and the initial state is chaotic; we suspect the combination of this with the discontinuity in \textit{accelerating shock} causes exploding gradients for which we found no sensible means of correcting. For this analysis, we only train with the other two training initial conditions, \textit{standing sine} and \textit{rarefaction}. To get a fair comparison, we evaluate the learned policies with the original Markovian reward function. This is pictured in Figure~\ref{fig:burgers:supervised}. The data here are averaged over the two initial conditions used during training, and only for the 0.1 second training duration.
Training against the true solution is unstable. Unlike the WENO solution, the shape of the true solution stays sharp throughout the episode. This leads to large misleading gradients. Training against the fixed version of the standard WENO solution fares better but improves much more slowly than with the original reward. It also does not generalize well to \textit{accelerating shock} as this type of physics is missing during training.

\subsection{Additional Action Plots}
\begin{figure}[h]
	\centering
	\includegraphics[width=\linewidth]{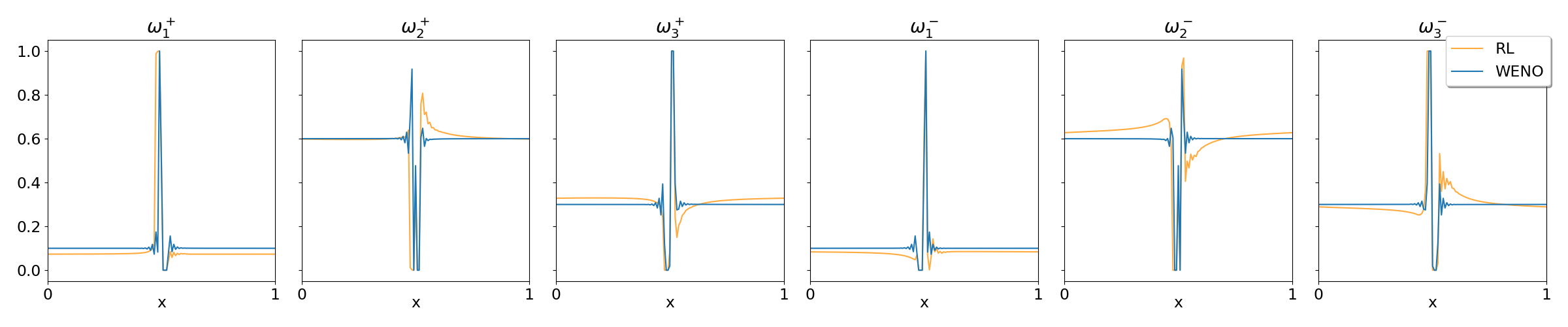}
	\vspace{-7mm}
	\caption{Comparison of actions on the \textit{standing sine} initial condition (Figure~\ref{fig:burgers:eval:sine}).}
	\label{fig:burgers:actions:sine}
\end{figure}

\begin{figure}[h]
	\centering
	\includegraphics[width=\linewidth]{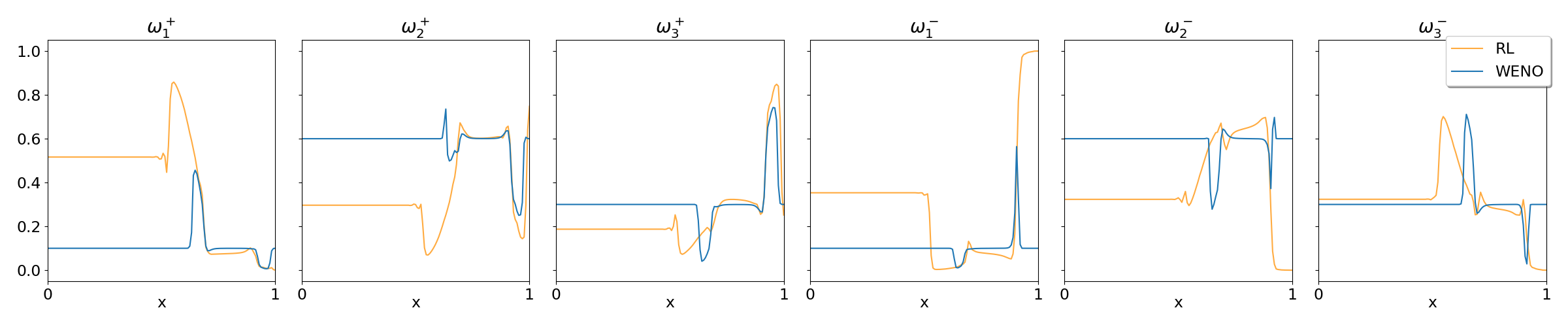}
	\vspace{-7mm}
	\caption{Comparison of actions on the \textit{rarefaction} initial condition 	(Figure~\ref{fig:burgers:eval:rarefaction}).}
	\label{fig:burgers:actions:rarefaction}
\end{figure}

\begin{figure}[h]
	\centering
	\includegraphics[width=\linewidth]{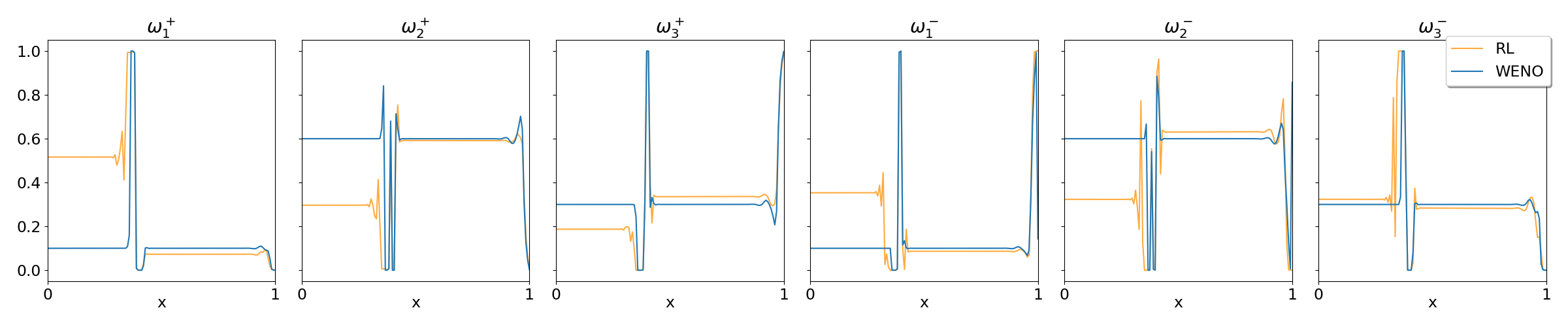}
	\vspace{-7mm}
	\caption{Comparison of actions on the \textit{accelerating shock} initial condition (Figure~\ref{fig:burgers:eval:accelshock}).}
	\label{fig:burgers:actions:accelshock}
\end{figure}

\begin{figure}[H]
	\centering
	\includegraphics[width=\linewidth]{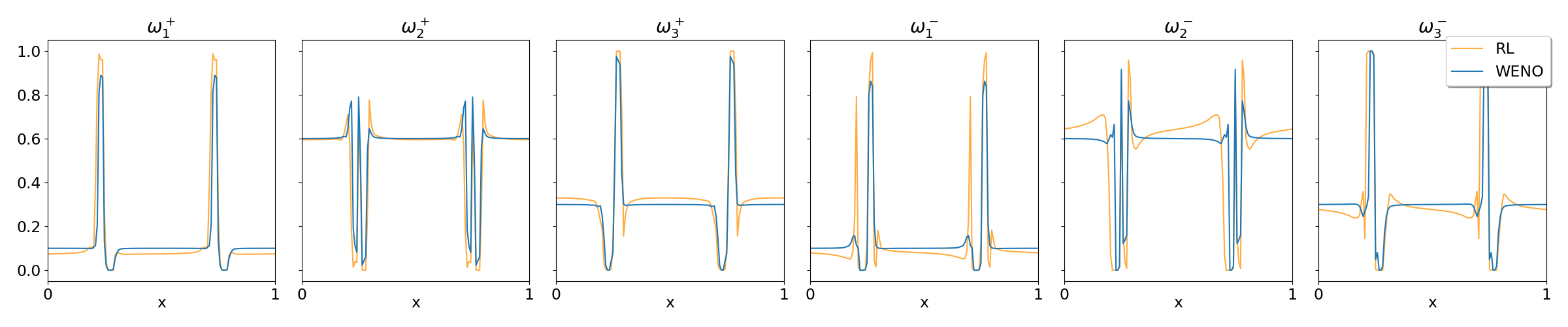}
	\vspace{-7mm}
	\caption{Comparison of actions on the \textit{double sine} initial condition (Figure~\ref{fig:burgers:eval:other}).}
	\label{fig:burgers:actions:other}
\end{figure}

\begin{figure}[H]
	\centering
	\includegraphics[width=\linewidth]{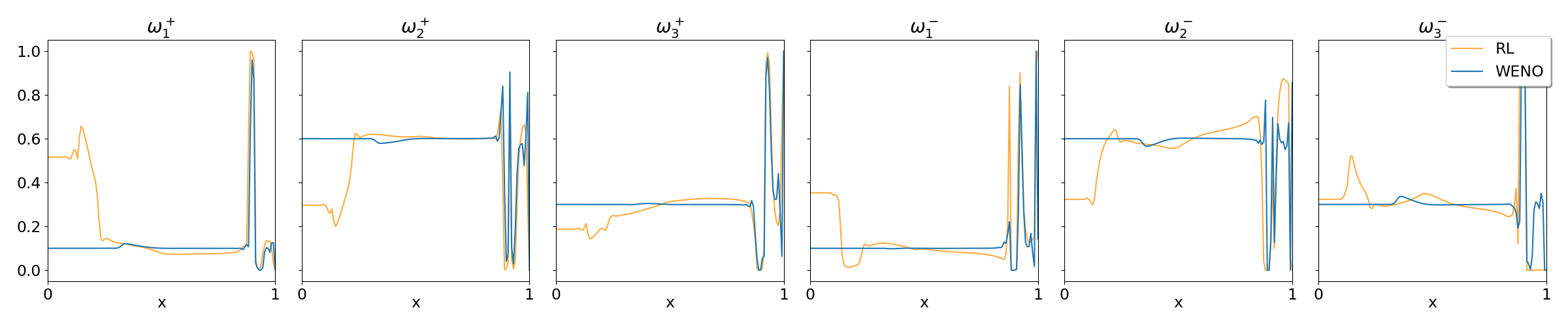}
	\vspace{-7mm}
	\caption{Comparison of actions on the \textit{gaussian} initial condition (Figure~\ref{fig:burgers:eval:gaussian}).}
	\label{fig:burgers:actions:gaussian}
\end{figure}

\begin{figure}[H]
	\centering
	\includegraphics[width=\linewidth]{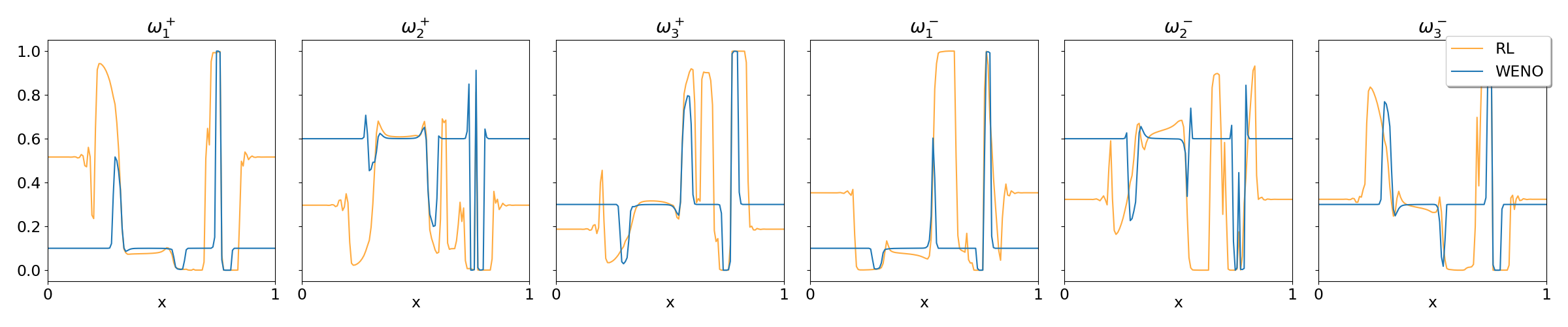}
	\vspace{-7mm}
	\caption{Comparison of actions on the \textit{tophat} initial condition (Figure~\ref{fig:burgers:eval:tophat}).}
	\label{fig:burgers:actions:tophat}
\end{figure}

In this section, we present the action values of the policy for all six initial conditions in Figure~\ref{fig:burgers:eval}, in addition to the actions for \textit{tophat} in Figure~\ref{fig:burgers:actions} above. These are shown in Figures~\ref{fig:burgers:actions:sine}-\ref{fig:burgers:actions:tophat}. The actions are from the second-to-last timestep of the episode. These actions show that while the policy frequently uses the same weights as the standard WENO scheme, it does not always do so. This difference is evidence that the policy learns its own understanding of the underlying physics instead of simply copying the WENO solution.


\pagebreak
\section{Addition Euler Equation Experiments}
\label{appendix:euler_plot}
\begin{figure}[ht]
	\centering
	\begin{subfigure}{1.0\linewidth}
	\begin{subfigure}{0.33\linewidth}
	\includegraphics[width=\linewidth]{Figures/euler/sod_rho}
\end{subfigure}
\begin{subfigure}{0.33\linewidth}
	\includegraphics[width=\linewidth]{Figures/euler/sod_u}
\end{subfigure}
\begin{subfigure}{0.33\linewidth}
	\includegraphics[width=\linewidth]{Figures/euler/sod_p}
\end{subfigure}
		\caption{Comparison of the state reached using the learned policy and the standard WENO scheme for the Sod initial condition.}
		\label{fig:euler:sod:action}
	\end{subfigure}
		\begin{subfigure}{1.0\linewidth}
		\includegraphics[width=\linewidth]{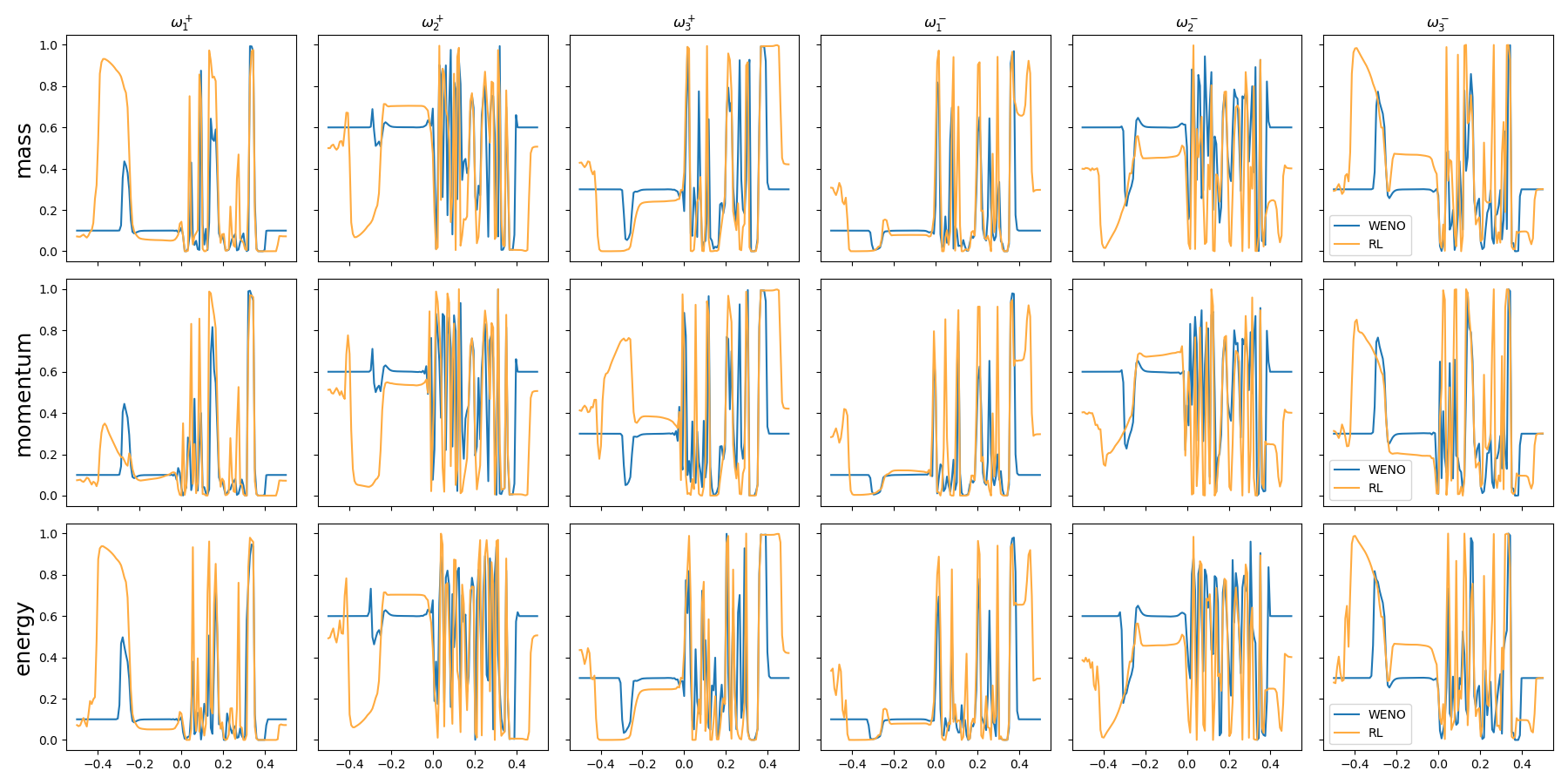}
		\caption{Final step actions for mass, momentum, and energy fluxes for the learned policy and the equivalent policy of the standard WENO scheme on the Sod initial condition}
		\label{fig:euler:sod_action}
	\end{subfigure}
  \caption{Test results on the Sod initial condition.}
  \label{fig:euler:sod}
\end{figure}

\begin{figure}[ht]
	\centering
	\begin{subfigure}{1.0\linewidth}
	\begin{subfigure}{0.33\linewidth}
	\includegraphics[width=\linewidth]{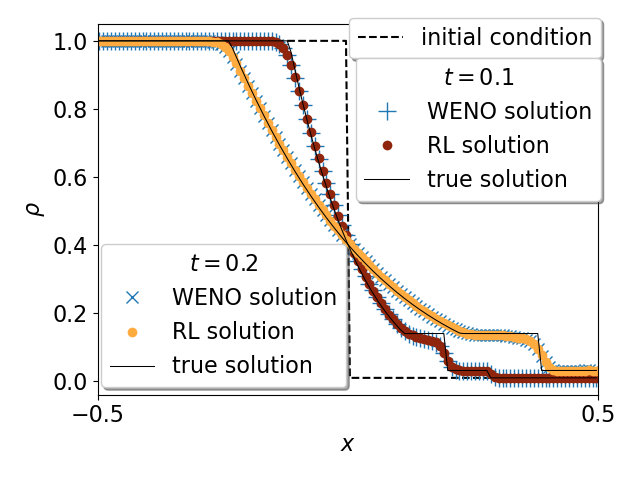}
\end{subfigure}
\begin{subfigure}{0.33\linewidth}
	\includegraphics[width=\linewidth]{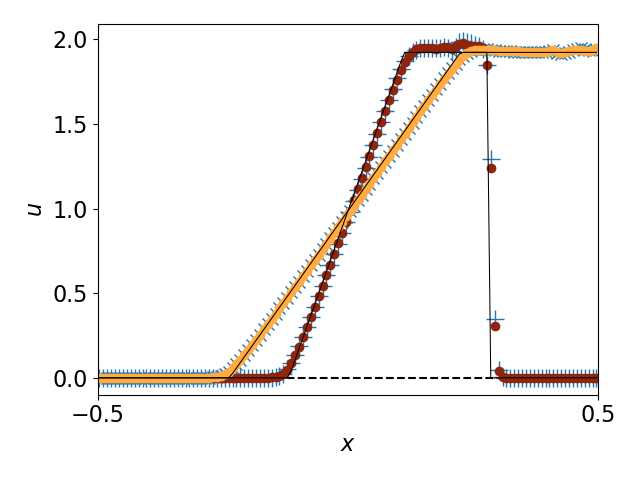}
\end{subfigure}
\begin{subfigure}{0.33\linewidth}
	\includegraphics[width=\linewidth]{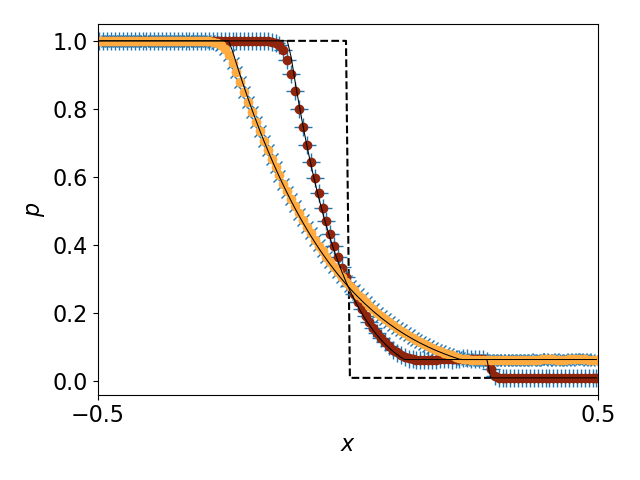}
\end{subfigure}
		\caption{Comparison of the state reached using the learned policy and the standard WENO scheme for the Sod2 initial condition.}
		\label{fig:euler:sod2:state}
	\end{subfigure}
	\begin{subfigure}{1.0\linewidth}
		\includegraphics[width=\linewidth]{Figures/euler/sod_final_actions.png}
		\caption{Final step actions for mass, momentum, and energy fluxes for the learned policy and the equivalent policy of the standard WENO scheme on the Sod2 initial condition}
		\label{fig:euler:sod2_action}
	\end{subfigure}
  \caption{Test results on the Sod2 initial condition.}
    \label{fig:euler:sod2}
\end{figure}

\begin{figure}[ht]
	\centering
	\begin{subfigure}{1.0\linewidth}
	\begin{subfigure}{0.33\linewidth}
	\includegraphics[width=\linewidth]{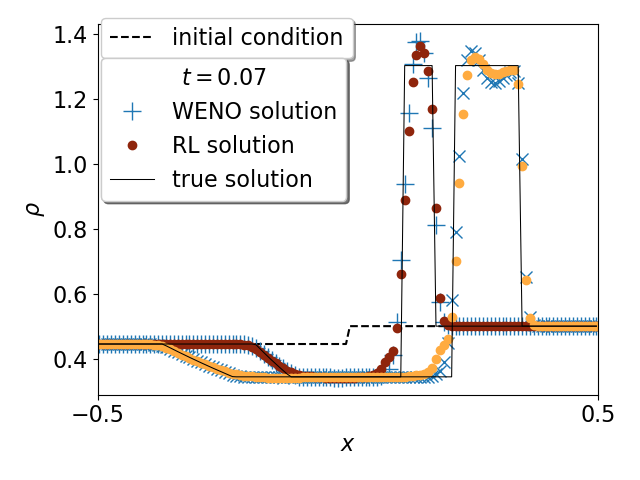}
\end{subfigure}
\begin{subfigure}{0.33\linewidth}
	\includegraphics[width=\linewidth]{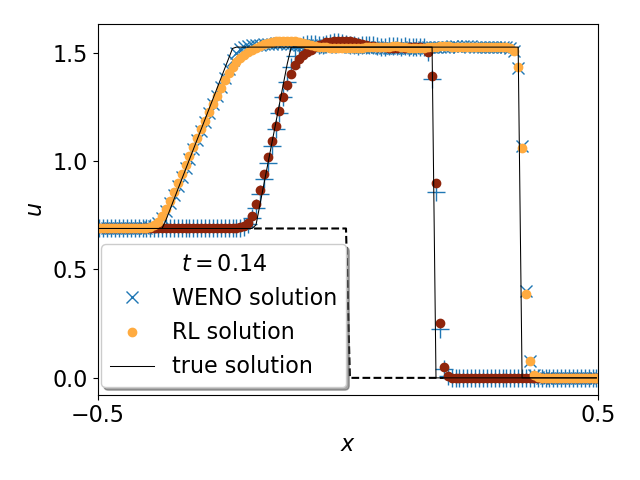}
\end{subfigure}
\begin{subfigure}{0.33\linewidth}
	\includegraphics[width=\linewidth]{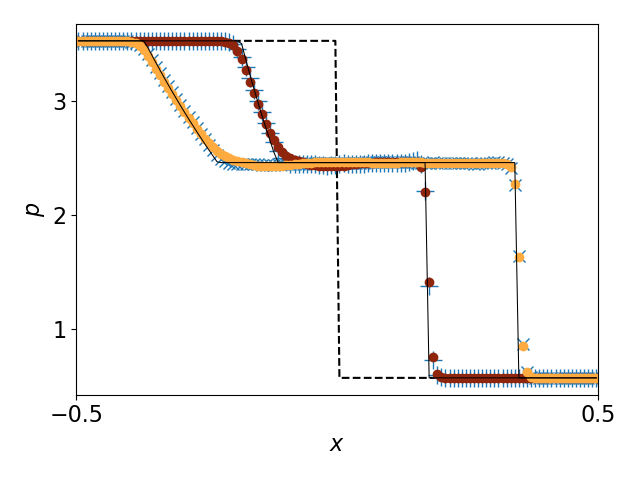}
\end{subfigure}
		\caption{Comparison of the state reached using the learned policy and the standard WENO scheme for the Lax initial condition.}
		\label{fig:euler:lax:state}
	\end{subfigure}
	\begin{subfigure}{1.0\linewidth}
		\includegraphics[width=\linewidth]{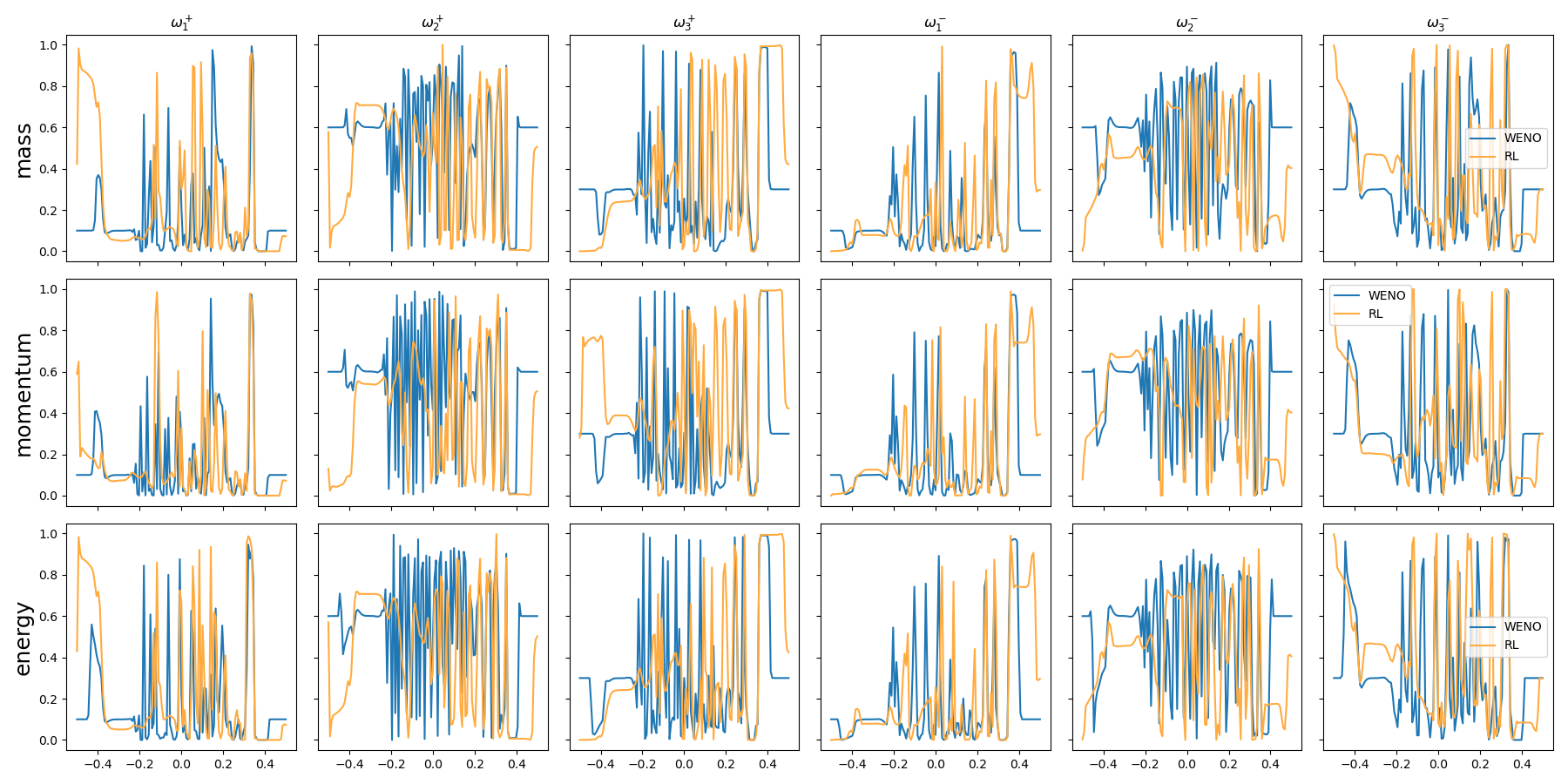}
		\caption{Final step actions for mass, momentum, and energy fluxes for the learned policy and the equivalent policy of the standard WENO scheme on the Lax initial condition}
		\label{fig:euler:lax_action}
	\end{subfigure}
  \caption{Test results on the Lax initial condition.}
    \label{fig:euler:lax}
\end{figure}

\begin{figure}[ht]
	\centering
	\begin{subfigure}{1.0\linewidth}
	\begin{subfigure}{0.33\linewidth}
	\includegraphics[width=\linewidth]{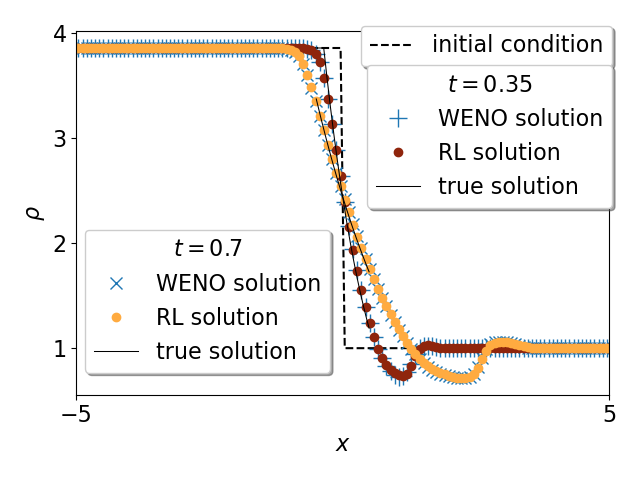}
\end{subfigure}
\begin{subfigure}{0.33\linewidth}
	\includegraphics[width=\linewidth]{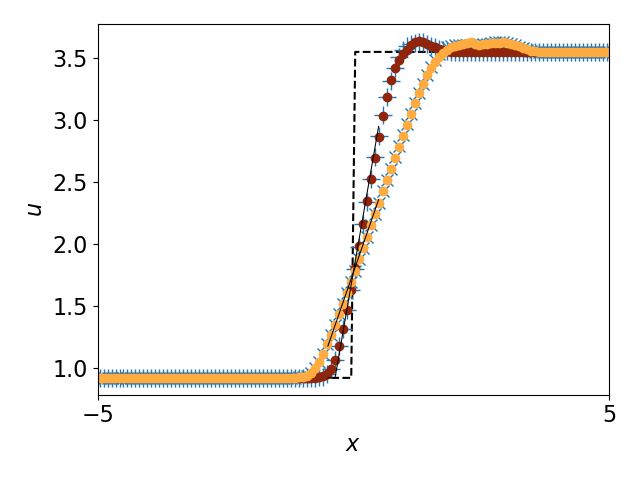}
\end{subfigure}
\begin{subfigure}{0.33\linewidth}
	\includegraphics[width=\linewidth]{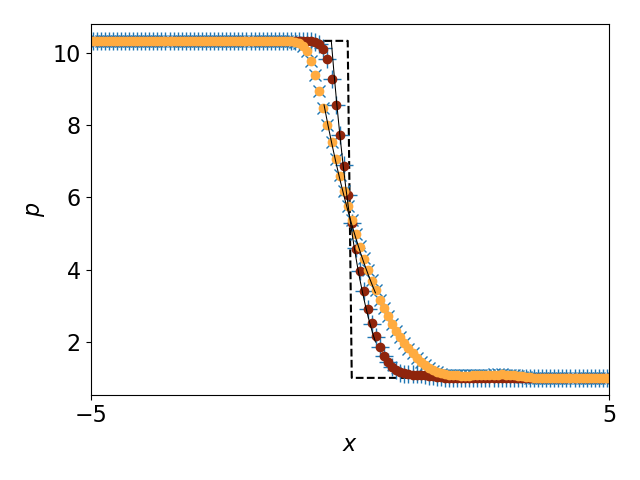}
\end{subfigure}
		\caption{Comparison of the state reached using the learned policy and the standard WENO scheme for the Sonic Rarefaction initial condition.}
		\label{fig:euler:sonic_rare:state}
	\end{subfigure}
	\begin{subfigure}{1.0\linewidth}
		\includegraphics[width=\linewidth]{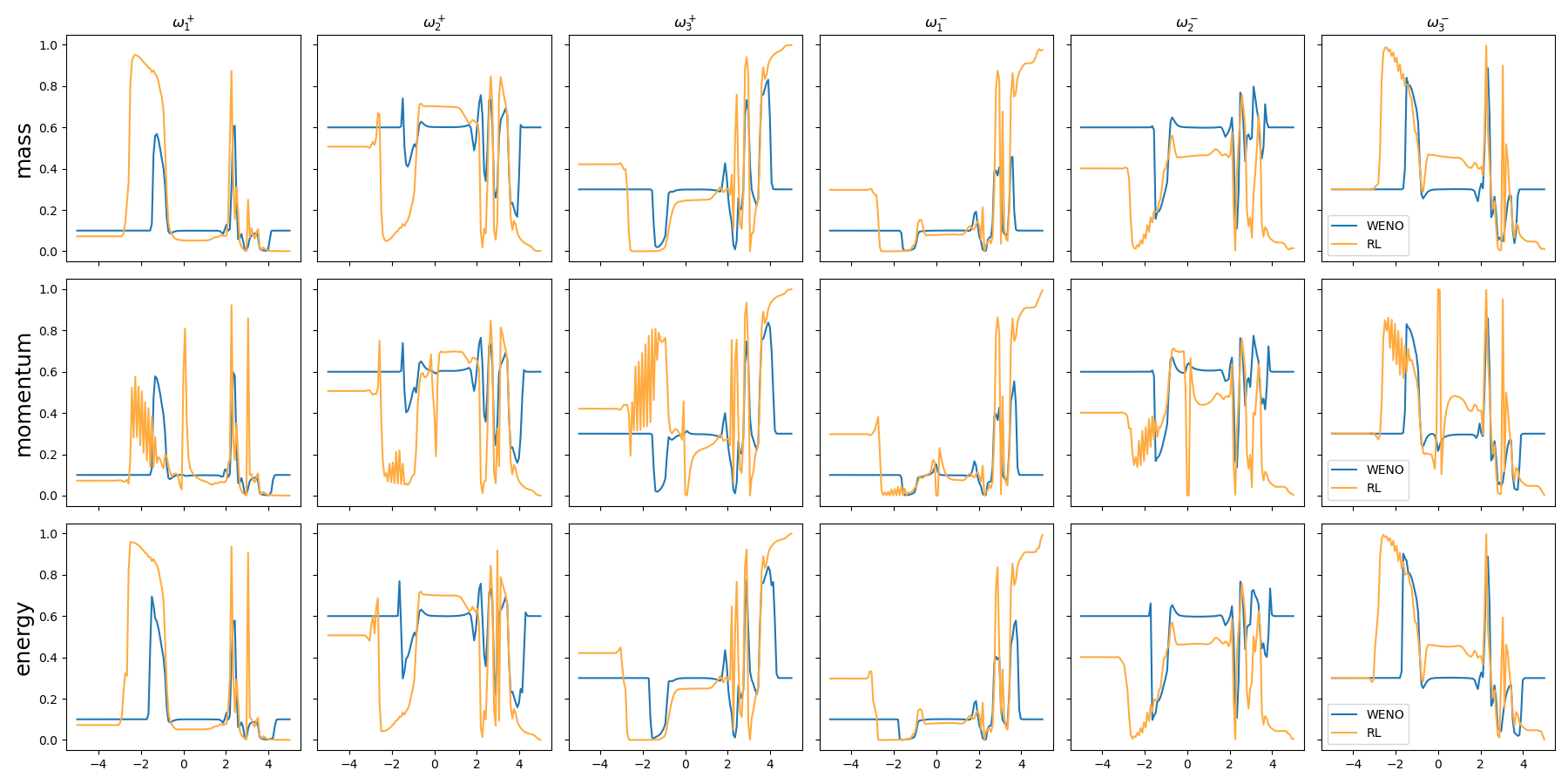}
		\caption{Final step actions for mass, momentum, and energy fluxes for the learned policy and the equivalent policy of the standard WENO scheme on the Sonic Rarefaction initial condition}
		\label{fig:euler:sonic_rare_action}
	\end{subfigure}
  \caption{Test results on the Sonic Rarefaction initial condition.}
  \label{fig:euler:sonic_rare}
\end{figure}

In this section, we present the state values and corresponding second-to-last actions for evaluating the learned policy on the Euler equations. These are shown in Figures~\ref{fig:euler:sod}-\ref{fig:euler:sonic_rare}.

\end{document}